\documentclass[notitlepage,12pt]{jedm}
\usepackage[table]{xcolor}
\usepackage{url}
\usepackage{hyperref}
\usepackage{graphicx}
\usepackage{amsmath}
\usepackage{amssymb}
\usepackage{array}
\usepackage{tikz}
\usetikzlibrary{calc}
\usetikzlibrary{angles, quotes} 
\hypersetup{
  colorlinks   = true, 
  urlcolor     = blue, 
  linkcolor    = blue, 
  citecolor   = blue 
}

\begin{document}

\title{Evaluating NLP Embedding Models for Handling Science-Specific Symbolic Expressions in Student Texts}
\date{} 


\newcommand{\authorFixedWidth}[1]{\parbox[t]{.25\textwidth}{\raggedright#1 \raisebox{0pt}[0pt][6pt]{}}}
\author{\authorFixedWidth{{\large Tom Bleckmann}\\IDMP - Physics Education Group, Leibniz University Hannover\\Hannover, Germany\\bleckmann@idmp.uni-hannover.de} \and \authorFixedWidth{{\large Paul Tschisgale}\\Leibniz Institute for Science and Mathematics Education\\Kiel, Germany\\tschisgale@leibniz-ipn.de}  }

\maketitle

\begin{abstract}
In recent years, natural language processing (NLP) has become integral to educational data mining, particularly in the analysis of student-generated language products. For research and assessment purposes, so-called embedding models are typically employed to generate numeric representations of text that capture its semantic content for use in subsequent quantitative analyses. Yet when it comes to science-related language, symbolic expressions such as equations and formulas introduce challenges that current embedding models struggle to address. Existing research studies and practical applications often either overlook these challenges or remove symbolic expressions altogether, potentially leading to biased research findings and diminished performance of practical applications. This study therefore explores how contemporary embedding models differ in their capability to process and interpret science-related symbolic expressions. To this end, various embedding models are evaluated using physics-specific symbolic expressions drawn from authentic student responses, with performance assessed via two approaches:\break 1) similarity-based analyses and 2) integration into a machine learning pipeline. Our findings reveal significant differences in model performance, with OpenAI’s \texttt{GPT-text-embedding-3-large} outperforming all other examined models, though its advantage over other models was moderate rather than decisive. Overall, this study underscores the importance for educational data mining researchers and practitioners of carefully selecting NLP embedding models when working with science-related language products that include symbolic expressions. The code and (partial) data are available at \url{https://doi.org/10.17605/OSF.IO/6XQVG}. \\ 

{\parindent0pt
\textbf{Keywords:} natural language processing, symbolic expressions, text embeddings, science language
}
\end{abstract}

\section{Introduction and Literature Review}

With the ongoing digital transformation of education, vast amounts of textual data are being generated that could provide valuable insights into students' learning and learning processes \cite{oecdOECDDigitalEducation2023}. However, qualitatively analyzing these large amounts of data requires significant time and effort from humans, which is why more scalable approaches must be employed \cite{chandrasekarMakingMostBig2024,jiangSupportingSerendipityOpportunities2021}. Recent advancements in artificial intelligence (AI), particularly in natural language processing (NLP), have substantially improved the scalable processing and analysis of textual data in educational data mining contexts \cite{ferreira-melloTextMiningEducation2019,liuQualitativeCodingGPT42025}, with NLP techniques now widely used to examine the large volumes of language products generated by students \cite{wuNaturalLanguageProcessing2024,zhaiApplyingMachineLearning2020}. 
For example, a systematic literature review conducted by \citeN{ferreira-melloWordsWisdomJourney2024} indicates that NLP embedding models have been the predominant approach in NLP-based learning analytics research between 2021 and 2023, generally outperforming other text analysis methods. The basic principle behind NLP embedding models is to transform words, sentences, or even entire documents into high-dimensional numerical vectors—known as \textit{embeddings}—that capture their semantic content \cite{qiuPretrainedModelsNatural2021}.
Embedding models typically rely on neural network architectures that are trained on large collections of text. They generate contextualized embeddings, meaning that they capture how the meaning of a word depends on its surrounding context \cite{ethayarajhHowContextualAre2019}. As a result, the generated embeddings position semantically similar inputs closer together in the embedding space, allowing for nuanced comparisons between textual elements. This embedding space typically has hundreds or even thousands of dimensions, capturing latent features that would be difficult to identify with traditional text analysis techniques. 
Embeddings are then typically utilized in further downstream tasks within educational data mining contexts, such as applying machine learning algorithms for classification or pattern recognition \cite{linComprehensiveSurveyDeep2025}. It is also important to note that such embedding models represent a fundamental component of contemporary large language models, such as ChatGPT \cite{brownLanguageModelsAre2020,petukhovaTextClusteringLarge2025}.

NLP embedding models have been successfully applied to analyze various types of student-generated language products, including for example students' argumentation \cite{martinExploringNewDepths2023}, concept maps \cite{bleckmannConceptMapsFormative2023}, constructed responses \cite{gombertCodingEnergyKnowledge2023}, problem solving \cite{tschisgaleIntegratingArtificialIntelligencebased2023}, and self-regulated learning \cite{borchersLargeLanguageModels2025}. While these NLP-based analyses of students' textual responses have yielded promising results, the language used in science-related contexts (e.g., in scientific arguments, explanations, and problem solutions) differs significantly from everyday language \cite{lemkeTalkingScienceLanguage1990}, particularly as it often includes symbolic expressions, such as formulas, equations, and standard scientific notations \cite{baralAutomatedScoringImagebased2023,brookesForceOntologyLanguage2009,ferreiraMathematicalLanguageProcessing2022,meadowsPhysNLULanguageResource2022}. These symbolic expressions may even serve as the core elements within language products, representing fundamental scientific concepts and relationships while enabling precise and formal communication of disciplinary knowledge \cite{treagustRoleSubmicroscopicSymbolic2003}. For example, in physics, when a student writes ``$E_{\text{total}}=\text{const.}$ holds'',  they succinctly encapsulate the conceptual idea of energy conservation; omitting the symbolic expression would strip the sentence of its essential meaning.
Although symbolic expressions are often intuitive for professionals in the sciences, their integration into natural language poses challenges for students \cite{delozanoLearningProblemsConcerning2002,maraisAreWeTaking2000}—and similarly for NLP embedding models. Challenges for such models in handling mixtures of text and symbolic expressions may arise in part because these combinations are rarely represented in their training data, and in part because contextual encoding of symbolic expressions may require explicit training \cite{ferreiraMathematicalLanguageProcessing2022}. 

Although some studies explicitly report difficulties in handling symbolic expressions within NLP-based analyses, many others may not even be aware of an issue. For example, \citeN{botelhoLeveragingNaturalLanguage2023} used supervised machine learning and identified a strong relationship between the presence of mathematical expressions in student responses and increased classification error rates. Similarly, \citeN{bleckmannFormativesAssessmentAuf2024} utilized supervised machine learning for automated assessment and observed lower predictive accuracies for textual elements containing symbolic expressions. Similarly in an unsupervised machine learning setting, \citeN{tschisgaleIntegratingArtificialIntelligencebased2023} found that textual elements containing symbolic expressions were largely grouped into one particular cluster, even though more conceptually appropriate clusters existed, suggesting that the utilized embedding model was unable to adequately capture the meaning of the symbolic expressions. To mitigate such issues, some studies have opted to exclude symbolic expressions from textual data \cite{bralinAnalysisStudentEssays2023,shangRepresentationExtractionPhysics2022}. However, both approaches—omitting symbolic expressions or neglecting the information they entail—risk losing potentially valuable information for assessment or research purposes. Since symbolic expressions, such as formulas and equations, often encapsulate key scientific concepts, disregarding them in textual data analysis may lead to systematic errors. In particular, students who more heavily rely on symbolic expressions in their responses may be disproportionately affected, resulting in biased assessments that systematically underestimate their performance or conceptual understanding. 

Thus, there is a clear need for better understanding to what extent contemporary NLP-based approaches relying on embedding models can effectively process symbolic expressions in science-related textual data. 
For this, an essential first step is to better understand how different embedding models handle textual data interwoven with symbolic expressions. Therefore, this study aims to compare the performance of various NLP embedding models in processing such symbolic expressions.
This study focuses on symbolic expressions in a physics context, given the field's strong reliance on the interplay between natural language and symbolic expressions \cite{wulffPhysicsLanguageLanguage2024}. Moreover, rather than using curated, well-formatted symbolic expressions, we focus on those found in authentic student responses. This approach particularly accounts for the notational variability in student responses (e.g., ``E=m*g*h'', ``E(pot)=gmh'', and ``Epot=Mg$\Delta$h'' all represent the same physics idea but differ in their notation), thereby addressing an additional potential challenge for NLP embedding models. 

Based on all these considerations, this study aims to answer the following research question:
\\ \\
To what extent can different NLP embedding models accurately process physics-specific symbolic expressions as they appear in authentic student responses?

\section{Methods}

To address the research question, we adopted two approaches: First, we conducted a similarity-based analysis to examine the extent to which selected embedding models relate physics-specific symbolic expressions with their literal translations and related physics concepts. In the second approach, we evaluated the performance of the same embedding models within a machine learning pipeline. 

For both approaches, we used textual data collected from German students, which included symbolic expressions and was generated in physics contexts. Although symbolic expressions such as formulas and equations are generally similar across languages, they were often embedded within German text rather than appearing in isolation. 
Therefore, the selection of embedding models investigated in this study, summarized in Table~\ref{tab2}, includes both German-specific and multilingual embedding models. This selection also comprises a mix of open-source and proprietary models, aiming to critically examine the frequently observed performance advantage of proprietary models over open-source alternatives across a range of physics-related tasks \cite{fengPHYSICSBenchmarkingFoundation2025,petukhovaTextClusteringLarge2025,wangSciBenchEvaluatingCollegelevel2024}. Furthermore, one of the selected embedding models is specifically optimized for science education \cite{latifGSciEdBERTContextualizedLLM2024}. 

\begin{table}[]
\caption{Overview of the NLP embedding models that are compared in this study.}\vspace*{1ex}
\label{tab2}
\centering
\begin{tabular}{|>{\raggedright\arraybackslash}p{6.5cm} 
                |>{\raggedright\arraybackslash}p{6.2cm} 
                |c|}
\hline
Model name                 & Short description                                           & Dimension \\
\hline
\texttt{German\_Semantic\_STS\_V2}  & German sentence-transformers model          & 1024                  \\
\hline
\texttt{paraphrase-multilingual- mpnet-base-v2} & Multilingual sentence-transformers model \cite{reimersSentenceBERTSentenceEmbeddings2019} & 768 \\
\hline
\texttt{Gbert-large}                & German BERT language model \cite{chanGermansNextLanguage2020} & 1024                  \\
\hline
\texttt{G-SciEdBERT}                & German BERT language model finetuned for science education \cite{latifGSciEdBERTContextualizedLLM2024}          & 768                   \\
\hline
\texttt{GPT-text-embedding-3-large} & Multilingual third-generation embedding model from OpenAI                & 3072                  \\
\hline
\texttt{GPT-text-embedding-ada-002} & Multilingual second-generation embedding model from OpenAI               & 1536 \\
\hline  
\end{tabular}%
\end{table}

\subsection{Evaluation of Embedding Model Performance via Similarities}
\subsubsection{Data Sources and Preparation} \label{subsec:Data}
In the first approach, we examined textual responses from German students to two physics tasks \cite{bleckmannFormativesAssessmentAuf2024,tschisgaleIntegratingArtificialIntelligencebased2023}.
In one task, students were given an incomplete concept map in which the concepts and arrows between concepts were predefined, and students had to label these arrows, representing the relations between concepts  \cite{bleckmannFormativesAssessmentAuf2024}. In the other task, students had to describe their approach to solving a physics problem, i.e., outlining the ideas they would use and the steps they would make when actually solving the problem \cite{tschisgaleIntegratingArtificialIntelligencebased2023}.
In both tasks, students' responses were visually checked by the authors for symbolic expressions, which were then extracted by hand without their surrounding contextual information (e.g., the sentence in which they were embedded). While contextualized embeddings are designed to leverage surrounding context, the present approach deliberately omits contextual information to establish a lower bound for NLP embedding models' capabilities to handle symbolic expressions. In practical applications where contextual information are available (e.g., entire sentences in which symbolic expressions are embedded), model performance would likely be higher. This design choice allows for a clearer assessment of embedding models' inherent capabilities in processing symbolic expressions when no contextual information is available. Consequently, the upcoming analyses focus on the pure symbolic expressions found in students' written responses. In this way, a corpus of $N = 100$ symbolic expressions was generated, with an equal distribution between the two task types: 50\% from the concept map task and 50\% from the problem-solving task.

To assess the extent to which different embedding models effectively process students' symbolic expressions $\text{SE}_i$, five categories of expression-text pairs $(\text{SE}_i,\cdot)$ were constructed:
\begin{enumerate}
    \item \textbf{Category LT:} Pairs $(\text{SE}_i,\text{LT}_i)$ consisting of literal translations $\text{LT}_i$ of symbolic expressions $\text{SE}_i$.
    \item \textbf{Category ILT:} Pairs $(\text{SE}_i,\text{ILT}_i)$ containing incorrect literal translations $\text{ILT}_i$ of symbolic expressions $\text{SE}_i$, created by modifying one or, in some cases, few key words from the corresponding correct literal translation $\text{LT}_i$.
    \item \textbf{Category RC:} Pairs $(\text{SE}_i,\text{RC}_i)$ incorporating the primary related concept $\text{RC}_i$ for each symbolic expression $\text{SE}_i$, provided that it could be determined in a straightforward and unambiguous manner.
    \item \textbf{Category IRC:} Pairs $(\text{SE}_i,\text{IRC}_i)$ including incorrect related concepts $\text{IRC}_i$ that are somewhat related to the correct related concept $\text{RC}_i$ but not directly associated with the symbolic expression $\text{SE}_i$ itself.
    \item \textbf{Category OT:} Pairs $(\text{SE}_i,\text{OT}_i)$ where $\text{OT}_i=\text{``apple pie recipe''}$, serving as an off-topic (OT) reference that was entirely unrelated to physics and hence also unrelated to all symbolic expressions $\text{SE}_i$.
\end{enumerate}

The authors of this study have formulated these expression-text pairs for each symbolic expression. Examples are provided in Table~\ref{tab1}.

\begin{table}[]
\caption{Three exemplary symbolic expressions (SE) from the data corpus, including their literal translations (LT), incorrect literal translations (ILT), related concepts (RC), incorrect related concepts (IRC), and off-topic statements (OT); all translated from German into English by the authors. For some symbolic expressions (as in example 3), we refrained from stating (incorrect) related concepts if no clear physics concept was identifiable or the decision for such concepts would have been highly ambiguous.}\vspace*{1ex}
\label{tab1}
\centering
\begin{tabular}{|p{1.5cm} | p{3.6cm} | p{3.6cm} | p{3.6cm} | }
\hline
Category & Example 1                         & Example 2                      & Example 3          \\
\hline
SE       & “m·g·h= 1/2·m·v²”                 & “V=AXT”                        & “v=sqrt(g*r)”      \\
\hline
LT &
  “mass times gravitational acceleration times height equals half mass times velocity squared” &
  “velocity equals acceleration times time” &
  “velocity equals square root of gravitational acceleration times radius” \\
  \hline
ILT &
  “mass times gravitational acceleration times height equals half mass times acceleration squared” &
  “velocity is constant” &
  “acceleration equals square root of gravitational acceleration times radius” \\
  \hline
RC       & “law of conservation of energy”   & “uniformly accelerated motion” & NA                 \\
\hline
IRC      & “law of conservation of momentum” & “uniform motion”               & NA                 \\
\hline
OT       & “apple pie recipe”                & “apple pie recipe”             & “apple pie recipe” \\
\hline
\end{tabular}%

\end{table}

\subsubsection{Computing Similarities of Pairs}
For each NLP embedding model $M$ (see Table~\ref{tab2}), all symbolic expressions $\text{SE}_i$ and their corresponding textual counterparts $\text{LT}_i, \text{ILT}_i, \text{RC}_i, \text{IRC}_i, \text{and } \text{OT}_i$ were mapped to their respective $d$-dimensional embedding representations: $\mathbf{e}_{\text{SE}_i}^M$, $\mathbf{e}_{\text{LT}_i}^M$, $\mathbf{e}_{\text{ILT}_i}^M$, $\mathbf{e}_{\text{RC}_i}^M$, $\mathbf{e}_{\text{IRC}_i}^M$, and  $\mathbf{e}_{\text{OT}_i}^M$. The embedding dimension $d$ depends on the chosen embedding model $M$ (see Table~\ref{tab2}).

\begin{figure}
\centering
\Description{A two-dimensional coordinate system is shown, with the horizontal axis labeled $d_1$ and the vertical axis labeled $d_2$. Four points are depicted, each connected to the origin by an arrow.

The first point is located in the first quadrant and labeled $\text{SE}_i$: “m·g·h= 1/2·m·v²”. The corresponding arrow from the origin is labeled $\mathbf{e}{\text{SE}_i}^M$.

Slightly below and to the right of this point, still in the first quadrant, is a second point labeled $\text{RC}_i$: “law of conservation of energy“. Its arrow from the origin is labeled $\mathbf{e}{\text{RC}_i}^M$. An angle $\alpha$ is drawn between the arrows to $\text{SE}_i$ and $\text{RC}_i$.

Below the $\text{RC}_i$ point, and again slightly to the right but still within the first quadrant near the $d_1$-axis, is the third point labeled $\text{IRC}_i$: “law of conservation of momentum”. The arrow to this point is labeled $\mathbf{e}{\text{IRC}_i}^M$. An angle $\beta$ is drawn between this arrow and the arrow to $\text{SE}_i$.

The fourth point is located in the middle of the second quadrant and labeled $\text{OT}_i$: “apple pie recipe”. The arrow from the origin to this point is labeled $\mathbf{e}{\text{OT}_i}^M$. An angle $\gamma$ is drawn between this arrow and the arrow to $\text{SE}_i$.

Overall, angle $\alpha$ is the smallest, $\beta$ is larger, and $\gamma$ is the largest.}    
\caption{Illustration of the expected embedding structure if a model $M$ is capable of accurately processing and interpreting symbolic expressions. Using Example 1 from Table~\ref{tab2}, the corresponding embedding vectors $\mathbf{e}_{}^M$ are shown in a simplified two-dimensional embedding space (for illustration purposes). The plot includes embedding vectors for the specific symbolic expression $\text{SE}_i$, its related concept $\text{RC}_i$, an incorrect related concept $\text{IRC}_i$, and an off-topic statement $\text{OT}_i$. The angles $\alpha$, $\beta$, and $\gamma$ between embedding vectors reflect semantic similarity. Specifically, if the embedding model $M$ can accurately process and interpret a symbolic expression $\text{SE}_i$, the angle $\alpha$ between its embedding vector $\mathbf{e}_{\text{SE}_i}^M$ and that of a related concept $\mathbf{e}_{\text{RC}_i}^M$ should be smaller than the angle $\beta$ to the embedding of an incorrect related concept  $\mathbf{e}_{\text{IRC}_i}^M$, and much smaller than the angle $\gamma$ to the embedding of an off-topic statement $\mathbf{e}^M_{\text{OT}_i}$. } \label{tikz_plot}
\begin{tikzpicture}[scale=1.5, every node/.style={font=\footnotesize}]
  \draw[->] (-3,0) -- (3,0) node[right] {$d_1$};
  \draw[->] (0,-0.5) -- (0,3) node[left] {$d_2$};

  \coordinate (RC) at (2,1.9);
  \coordinate (IRC) at (2.5,0.7);
  \coordinate (SE) at (1.5,2.5);
  \coordinate (OT) at (-2,0.8);
  \coordinate (O) at (0,0);

  \draw[->, thick, black] (0,0) -- (RC) node[pos=0.7, below right, yshift=5pt] {$\mathbf{e}_{\text{RC}_i}^M$};
  \draw[->, thick, black] (0,0) -- (IRC) node[pos=0.7, below] {$\mathbf{e}_{\text{IRC}_i}^M$};
  \draw[->, thick, black] (0,0) -- (SE) node[pos=0.7, above left, xshift=2pt] {$\mathbf{e}_{\text{SE}_i}^M$};
  \draw[->, thick, black] (0,0) -- (OT) node[pos=0.6, above] {$\mathbf{e}_{\text{OT}_i}^M$};
  
  \filldraw[black] (RC) circle (0.8pt);
  \filldraw[black] (IRC) circle (0.8pt);
  \filldraw[black] (SE) circle (0.8pt);
  \filldraw[black] (OT) circle (0.8pt);
  \node[above right] at (RC) {$\text{RC}_i$: “law of conservation of energy“};
  \node[above right, xshift=-25pt] at (IRC) {$\text{IRC}_i$: “law of conservation of momentum“};
  \node[above right] at (SE) {$\text{SE}_i$: “m·g·h= 1/2·m·v²”};
  \node[above left, xshift=20pt] at (OT) {$\text{OT}_i$: “apple pie recipe“};

  \draw pic[draw=black, semithick, angle eccentricity=1.3, angle radius=0.8cm] {angle=IRC--O--SE};
  \node at ($(O)+(115:0.55cm)$) {$\gamma$};
  \draw pic["$\alpha$", draw=black, semithick, angle eccentricity=1.15, angle radius=1.3cm] {angle=RC--O--SE};
  \draw pic[draw=black, semithick, angle eccentricity=1.3, angle radius=0.6cm] {angle=SE--O--OT};
  \node at ($(O)+(30:0.7cm)$) {$\beta$};

\end{tikzpicture}
\end{figure}
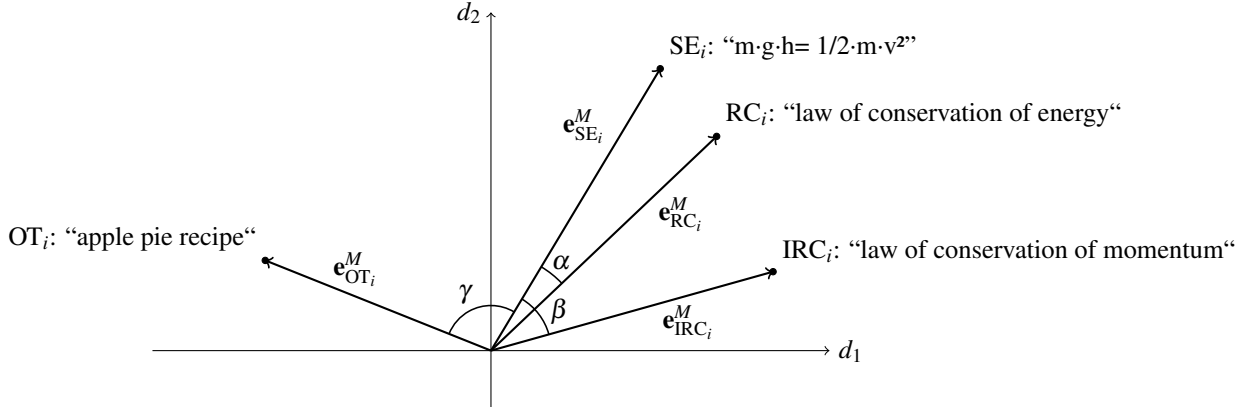

To assess the capability of different embedding models to capture the semantic meaning of symbolic expressions $\text{SE}_i$, we computed a similarity measure for each expression-text pair $(\text{SE}_i,\cdot)$ based on their respective model-dependent embeddings. Intuitively, two embeddings (which are simply high-dimensional vectors) from the same embedding model $M$ are considered more similar if the angle between them is small. This is because embedding models are trained to place semantically similar terms closer together in embedding space, which naturally results in smaller angles between their corresponding embedding vectors. This idea is illustrated and exemplified in Fig.~\ref{tikz_plot}.

A widely used metric for quantifying this similarity in high-dimensional embedding spaces is \textit{cosine similarity}, which is basically the cosine of the angle between two embeddings \cite{hirschleDeepNaturalLanguage2022}. 
Specifically, given two embeddings $\mathbf{e}_{A}^M= (e_{A,1}^M, \dots,e_{A,d}^M)$, $\mathbf{e}_{B}^M=(e_{B,1}^M, \dots,e_{B,d}^M)$, generated by the embedding model $M$ and serving as semantic representations of the textual statements “\textit{A}” and “\textit{B}”, the cosine similarity $S_{\cos}$ is given by
\begin{equation*}
    S_{\cos}(\mathbf{e}_A^M,\mathbf{e}_B^M)  =  
    \cos\left(\measuredangle(\mathbf{e}_A^M,\mathbf{e}_B^M)\right) = 
    \frac{\langle\mathbf{e}_A^M, \mathbf{e}_B^M\rangle}{\|\mathbf{e}_A^M\|\cdot \|\mathbf{e}_B^M \|} \in [-1,1],
\end{equation*}
where $\measuredangle(\mathbf{e}_A^M,\mathbf{e}_B^M)$ denotes the angle between the embedding vectors, and $\langle\mathbf{e}_A^M, \mathbf{e}_B^M\rangle$ is the scalar product of the embedding vectors given by
\begin{equation*}
    \langle\mathbf{e}_A^M, \mathbf{e}_B^M\rangle = \sum_{k=1}^{d} e_{A,k}^M\cdot e_{B,k}^M
\end{equation*}
and $\|\cdot\|=\sqrt{\langle\cdot,\cdot\rangle}$ is the Euclidean norm.

A cosine similarity of $1$ indicates that two embeddings point in the same direction within the embedding space, representing maximum similarity (e.g., synonymous terms), while a value of $-1$ indicates that they point in opposite directions, suggesting strong dissimilarity or antonymy. A value of $0$ implies that the embeddings are orthogonal, suggesting that there is no semantic relationship between the corresponding textual statements (which is the idea behind the off-topic category).

\subsubsection{Statistical Analyses}
We aimed to investigate how embedding model-dependent cosine similarity values can be used to distinguish between correct and incorrect expression-text pairs. In other words, we examined whether for specific embedding models higher similarity scores were assigned to correct pairs (e.g., ``V=AXT'' as symbolic expression and ``velocity equals acceleration times time'' as a correct literal translation) and lower scores to incorrect pairs (e.g., ``V=AXT'' and ``velocity is constant'' as an incorrect literal translation), thereby indicating an embedding model’s capability to capture meaningful semantic relationships with regard to symbolic expressions. More precisely, two complementary analyses were performed.

The first analysis utilized receiver operating characteristic (ROC) curves, as they allow quantifying the extent to which higher similarity scores are more strongly associated with correct pairs and lower scores with incorrect pairs, while specifically focusing on the separability of these two categories. In particular, ROC curves assess how effectively different similarity score thresholds can distinguish between correct and incorrect expression-text pairs (i.e., LT vs. ILT, and RC vs. IRC) \cite{brownReceiverOperatingCharacteristics2006}. In our context, a ROC curve depicts the true positive rate against the false positive rate across various thresholds of cosine similarity $S_{\cos}$.
Values exceeding the threshold are classified as correct (LT or RC), while values below the threshold are classified as incorrect (ILT or IRC). This method provides a visual impression of an embedding model’s capability to differentiate between correct and incorrect cases solely based on cosine similarity values. To quantify this capability, area under the curve ($\text{AUC}\in[0,1]$) is computed for each embedding model, with $\text{AUC} = 1$ representing a perfect classifier, $\text{AUC}=0.5$ indicating random guessing, and $\text{AUC}=0$ indicating a classifier that systematically misclassifies all instances.

The second analysis focused on individual symbolic expressions rather than the whole distribution of similarity values. If an embedding model $M$ effectively captures the semantics of symbolic expressions, we expect that the similarity between the embedding of a given symbolic expression $\text{SE}_i$ and the embedding of its corresponding literal translation $\text{LT}_i$ (or related concept $\text{RC}_i$) should be greater than its similarity to the embedding of an incorrect literal translation $\text{ILT}_i$ (or an incorrect related concept $\text{IRC}_i$), i.e., in more mathematical terms:
\begin{subequations}
\begin{align}
S_{\cos}(\mathbf{e}_{\text{SE}_i}^M, \mathbf{e}_{\text{LT}_i}^M ) &\overset{!}{>} S_{\cos}(\mathbf{e}_{\text{SE}_i}^M, \mathbf{e}_{\text{ILT}_i}^M ) \label{eq:sim1} \\
S_{\cos}(\mathbf{e}_{\text{SE}_i}^M, \mathbf{e}_{\text{RC}_i}^M ) &\overset{!}{>} S_{\cos}(\mathbf{e}_{\text{SE}_i}^M, \mathbf{e}_{\text{IRC}_i}^M ), \label{eq:sim2}
\end{align}
\end{subequations}
or rewritten in terms of the difference $\Delta S^M_{\cos}$ in model-dependent cosine similarity values:
\begin{subequations}
\begin{align}
\Delta S^M_{\cos}(\text{LT vs. ILT}) &:=S_{\cos}(\mathbf{e}_{\text{SE}_i}^M, \mathbf{e}_{\text{LT}_i}^M ) - S_{\cos}(\mathbf{e}_{\text{SE}_i}^M, \mathbf{e}_{\text{ILT}_i}^M )\overset{!}{>}0 \label{eq:sim3} \\
\Delta S^M_{\cos}(\text{RC vs. IRC}) &:=S_{\cos}(\mathbf{e}_{\text{SE}_i}^M, \mathbf{e}_{\text{RC}_i}^M ) - S_{\cos}(\mathbf{e}_{\text{SE}_i}^M, \mathbf{e}_{\text{IRC}_i}^M )\overset{!}{>}\textbf{}0, \label{eq:sim4}
\end{align}
\end{subequations}

A nonparametric statistical test that can assess whether similarity values of correct expression-text pairs tend to be greater than similarity values of corresponding incorrect expression-text pairs is the paired one-sided Wilcoxon signed-rank test \cite{woolsonWilcoxonSignedrankTest2005}. This test is a paired difference test, meaning it evaluates the distribution of differences by subtracting the similarity values of two paired categories associated with the same symbolic expression (see Eq.~(\ref{eq:sim3}) and Eq.~(\ref{eq:sim4})). 
The test then assesses whether the distribution of these differences is symmetric about zero (null hypothesis). We adopt a one-sided alternative hypothesis, where a significant result indicates that similarity values associated with correct categories (LT, RC) are systematically higher than those of their corresponding incorrect categories (ILT, IRC). Furthermore, we computed the matched-pairs rank biserial correlation coefficient as a measure of effect size \cite{kerbySimpleDifferenceFormula2014,kingStatisticalReasoningBehavioral2011}.

\subsection{Evaluation of Embedding Models within a Machine Learning Pipeline}
The second approach aimed to examine how the choice of a particular embedding model affects the performance of a supervised machine learning classifier that was trained on the respective model-specific embeddings. 
For this purpose, we used data from the physics concept mapping task, where students labeled predefined connections between concepts (cf. section \ref{subsec:Data}).

The concept maps resulted in a dataset of $N=3322$ textual elements or so called propositions (concept A -- linking statement -- concept B), the majority of which were purely textual ($83\%$; e.g., "velocity" -- "increases with" -- "free fall"), while a subset incorporated symbolic expressions to varying degrees ($17\%$; e.g., "force" -- "F = m*a" -- "acceleration"). Each proposition was evaluated for its correctness and level of detail using a rating scheme consisting of four categories (wrong answers, superficial answers, simple but more directed answers, detailed answers). 

In the original study \cite{bleckmannConceptMapsFormative2023}, after the propositions were preprocessed, different machine learning pipelines (consisting of an NLP embedding model and a machine learning classifier) were trained. More precisely,the \textit{German\_Semantic\_STS\_V2} embedding model was used to transform the propositions that were available in natural language into sentence embeddings. Based on those embeddings, different classifiers were trained, among them a support vector machine (SVM) classifier \cite{bishopPatternRecognitionMachine2006}. The aim was to find the classifier that could best automatically classify the propositions according to the four rating categories. All propositions were also coded by human raters to establish a gold standard. To assess the performance of trained classifiers, Cohen’s $\kappa$ was computed to quantify the agreement between a classifier’s predictions and the true category labels, while adjusting for chance agreement \cite{cohenCoefficientAgreementNominal1960}. During model training, the data was split into a training, validation and test data sets using a stratified cross-validation strategy involving hyperparameter optimization. For more information on model training, we refer to \citeN{bleckmannConceptMapsFormative2023}. Overall, it was found that a support vector machine (SVM) classifier provided the best Cohen’s $\kappa$ on the dataset.

Whereas the original study fixed the NLP embedding model and varied the machine learning classifiers, the present study takes the opposite approach: the classifier is fixed (using an SVM), while the NLP embedding models are varied. In other words, this study investigates how different NLP embedding models (see Table~\ref{tab2}) influence the classification performance of the SVM, as measured by Cohen’s $\kappa$. 


\section{Results}

\subsection{Descriptive Results}
\begin{figure}
\centering
\Description{The figure contains six subplots arranged in a 3×2 grid. Each subplot displays five vertical boxplots. The axes are identical across all subplots: the vertical axis ranges from 0 to 1 and is labeled “cosine similarity $S_\text{cos}$”, while the horizontal axis has five categories labeled “LT”, “ILT”, “RC”, “IRC”, and “OT” from left to right.

The subplot titles, ordered from left to right and top to bottom, are: German_Semantic_STS_V2, paraphrase-multilingual-mpnet-base-v2, Gbert-large, G-SciEdBERT, GPT-text-embedding-3-large, and GPT-text-embedding-ada-002.

While the exact positions and shapes of the boxplots vary between subplots, a consistent trend is visible: cosine similarity generally decreases slightly from left to right across the categories in each subplot.}  
\includegraphics[width=0.8\textwidth]{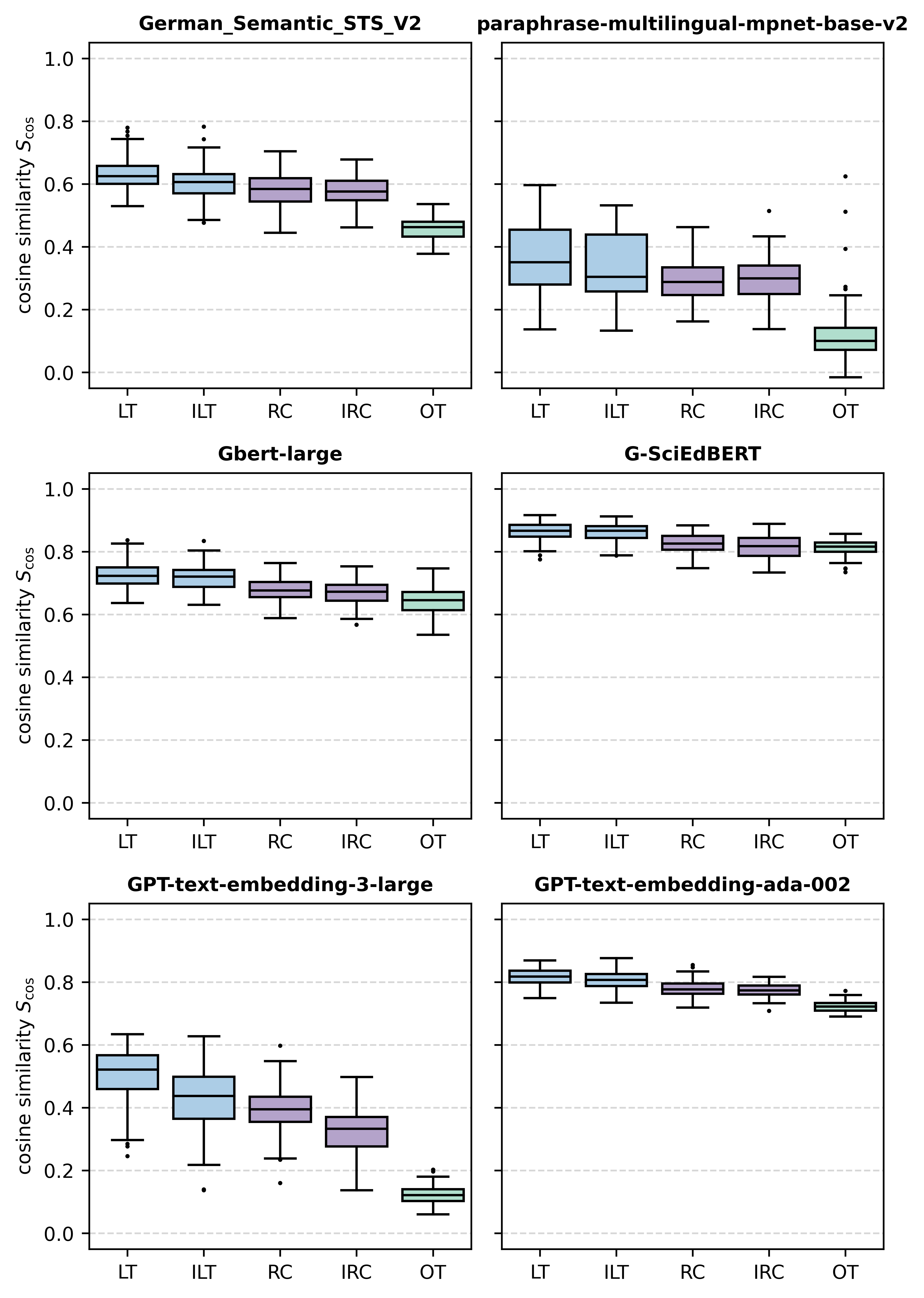}
\caption{Boxplot visualization of cosine similarity values across embedding models. For each embedding model $M$, there are five boxplots representing the distributions of similarity values $S_{\cos}(\mathbf{e}_{\text{SE}_i}^M, \mathbf{e}_{\text{C}_i}^M )$ for expression-text pairs of the five categories $C\in\{\text{LT}, \text{ILT}, \text{RC}, \text{IRC}, \text{OT}\}$ with LT: literal translations, ILT: incorrect literal translations, RC: related concepts, IRC: incorrect related concepts, OT: off-topic.} \label{fig1}
\end{figure}

The distributions of computed similarity values across embedding models are visualized in Fig.~\ref{fig1}, where each embedding model is represented by multiple boxplots corresponding to the five categories exemplified in Table~\ref{tab1}. On average, the cosine similarity between symbolic expressions and their correct literal translations or related concepts seems to be higher, albeit sometimes only slightly, than the similarity with incorrect literal translations or related concepts, respectively. This may be regarded as a first indicator that the embedding models are, to some extent, capable of interpreting symbolic expressions meaningfully. Interestingly, only the three models \texttt{German\_Semantics\_STS\_V2}, \texttt{paraphrase-multilingual-mpnet\-base-v2}, and \texttt{GPT-text-embedding-3-large} assign noticeably lower similarity scores to the off-topic statement (OT) compared to the more meaningful comparisons. Among these, only the latter two assign values close to zero, suggesting that they successfully detect a lack of semantic similarity between the symbolic expressions and the off-topic statement “apple pie recipe”.

\subsection{Evaluation of Embedding Model Performance via Similarities}
\label{sec:3.1}
We aimed to investigate how embedding model-dependent cosine similarity values can be used to distinguish between correct and incorrect expression-text pairs. To this end, two complementary analysis were conducted.

\begin{figure}[]
\centering
\Description{The figure contains two subplots. The left subplot is titled "ROC Curve: Correct versus Incorrect Literal Translations" and the right subplot is titled "ROC Curve: Correct versus Incorrect Related Concept".

In both subplots, the horizontal axis is labeled "false positive rate" and ranges from 0 to 1, while the vertical axis is labeled "true positive rate" and also ranges from 0 to 1. A black dashed diagonal line runs from (0,0) to (1,1). Six colored curves, slightly serrated, extend from the point (0,0) to the point (1,1). Most curves lie close to but slightly above the black dashed line. The yellow curve, corresponding to the "GPT-text-embedding-3-large" model, differs notably from the others by showing a strongly concave shape in both subplots.

Both subplots include a legend linking the colored curves to the embedding models, along with the corresponding AUC values.

In the left subplot, AUC values are as follows: GPT-text-embedding-3-large = 0.73, German_Semantic_STS_V2 = 0.64, GPT-text-embedding-ada-002 = 0.61, paraphrase-multilingual-mpnet-base-v2 = 0.57, Gbert-large = 0.53, G-SciEdBERT = 0.50.

In the right subplot, AUC values are as follows: GPT-text-embedding-3-large = 0.74, G-SciEdBERT = 0.58, Gbert-large = 0.56, GPT-text-embedding-ada-002 = 0.55, German_Semantic_STS_V2 = 0.52, paraphrase-multilingual-mpnet-base-v2 = 0.49.}  
\includegraphics[width=0.9\textwidth]{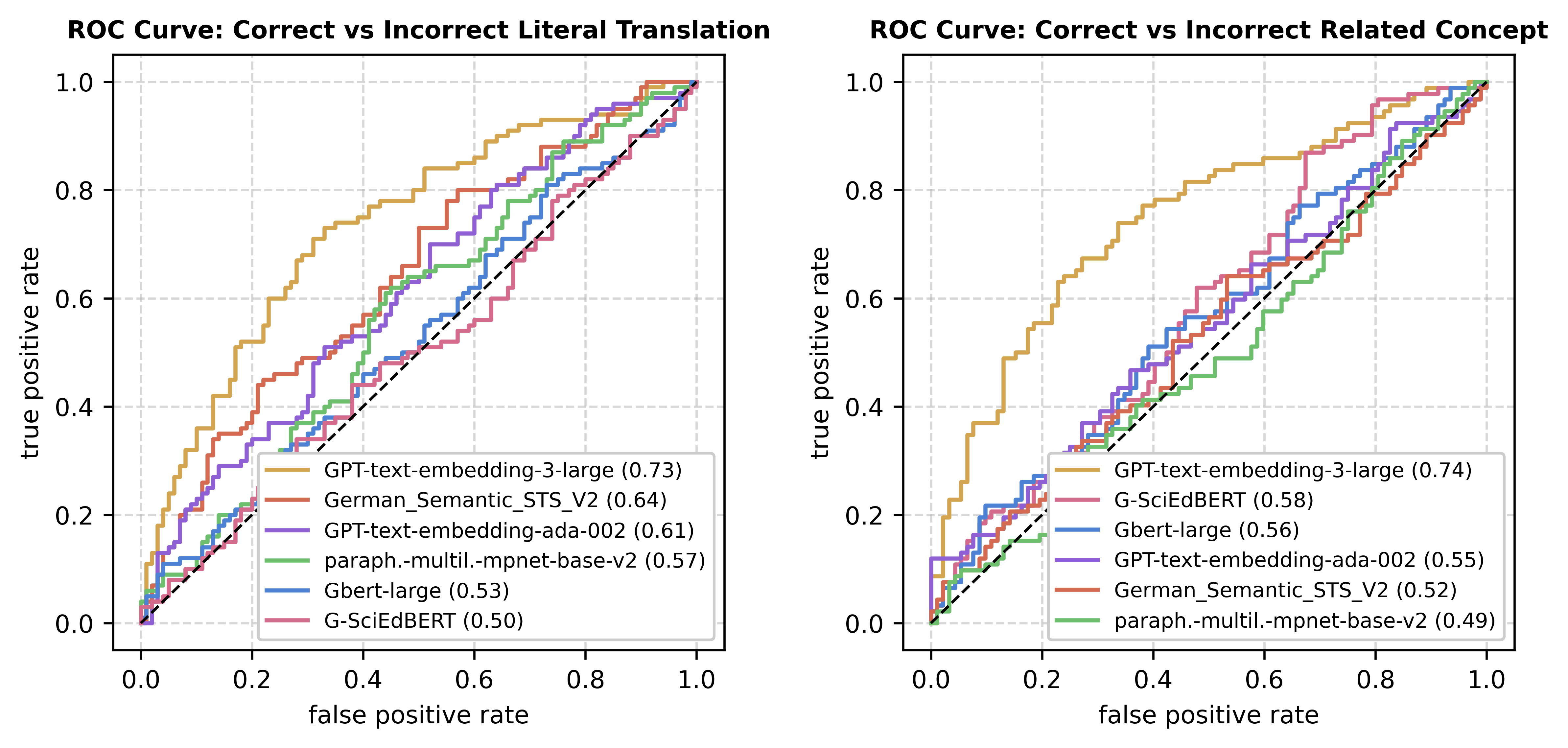}
\caption{Receiver Operating Characteristic (ROC) curves illustrating the capability of different embedding models to distinguish between correct and incorrect categories of expression-text pairs (left: LT vs. ILT; right: RC vs. IRC) based solely on model-dependent cosine similarity values. Curves closer to the top-left corner indicate better classification performance. The corresponding AUC values, ranked from highest to lowest, are displayed in the legend next to each embedding model.} \label{fig:ROC}
\end{figure}

The first analysis involved generating receiver operating characteristic (ROC) curves (see Fig.~\ref{fig:ROC}). Moreover, the area under curve (AUC) was computed for each embedding model as a measure of separability of correct and incorrect expression-text pairs. This analysis revealed that, for distinguishing correct from incorrect literal translations (LT vs. ILT; see Fig.~\ref{fig:ROC}, left), \texttt{GPT-text-embedding-3-large} is the most effective embedding model\break ($\text{AUC}~=~0.73$), although the AUC value only indicates moderate separability. Both \texttt{German\_ Semantic\_STS\_VS} ($\text{AUC}=0.64$) and \texttt{GPT-text-embedding-ada-002} ($\text{AUC}=0.61$) perform only slightly above random guessing. The remaining models performed close to random guessing ($\text{AUC}=0.57/0.53/0.50$).

For distinguishing correct from incorrect related concepts (RC vs. IRC; see Fig.~\ref{fig:ROC}, right), \texttt{GPT-text-embedding-3\allowbreak -large} again outperformed the other models ($\text{AUC}=0.74$). In particular, the remaining models do not demonstrate substantial improvement over random guessing ($\text{AUC}=0.49, \dots, 0.58$).

\begin{figure}
    \centering
    \Description{A green semicircle and a yellow right angle next to each other, with the words "Example text" between them.}  
    \includegraphics[width=1.00\linewidth]{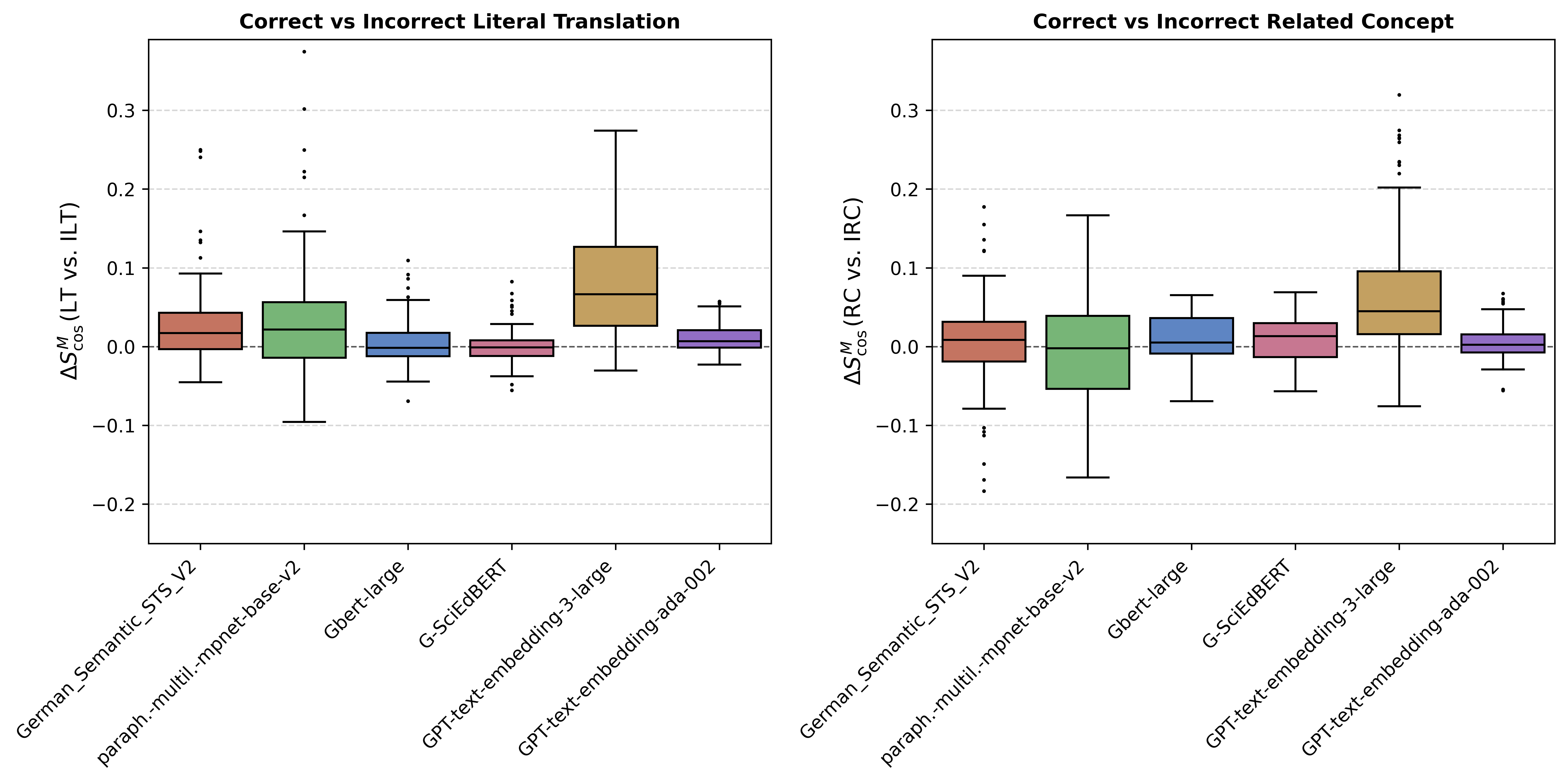}
    \caption{Boxplot visualization of the distribution of cosine similarity differences $\Delta S_{\cos}^M$ across the evaluated embedding models $M$, as defined in Eq.~(\ref{eq:sim3}) and Eq.~(\ref{eq:sim4}). Left: correct literal translations (LT) vs. incorrect literal translations (ILT). Right: Correct related concepts (RC) vs. incorrect related concepts (IRC). Better embedding model performance in handling symbolic expressions is indicated by a greater portion of the distribution lying above the horizontal zero line.}
    \label{fig:diff_distribution}
\end{figure}

The second analysis focused on examining similarity values at the level of individual expres-\allowbreak sion-text pairs by evaluating the differences in cosine similarity across embedding models for correct vs. incorrect literal translations (LT vs. ILT; see Eq.~(\ref{eq:sim3})) and correct vs. incorrect relate concepts (RC vs. IRC; see Eq.~(\ref{eq:sim4})). These embedding model-dependent distributions are visualized via boxplots in Fig.~\ref{fig:diff_distribution}. These boxplots already reveal that there are overall differences between the embedding models, and in particular, that some embedding models appear to struggle with reliably distinguishing between correct and incorrect literal translations as well as between correct and incorrect related concepts.

Differences between embedding models based on those distributions were further quantified: The results presented in Table~\ref{tab3} show the proportion of instances in which the similarity between a symbolic expression $\text{SE}_i$ and the correct literal translation  $\text{LT}_i$ (or correct related concept  $\text{RC}_i$) exceeds the expression's similarity with the corresponding incorrect literal translation $\text{ILT}_i$ (or incorrect related concept  $\text{IRC}_i$), as formally represented in Equations~(\ref{eq:sim1}) and (\ref{eq:sim2}). To assess statistical significance, we conducted a paired one-sided Wilcoxon signed-rank test and computed the matched-pairs rank biserial correlation coefficient as a measure of effect size.

The findings demonstrate that the \texttt{GPT-text-embedding-3-large} model consistently outperforms all other embedding models in both the LT vs. ILT (prop. $=91\%$, $d=0.95$) and RC vs. IRC comparison (prop. $=83\%$, $d=0.82$). However, the performance ranking of the remaining models depends on the comparison. In the case of literal translations (LT~vs.~ILT), the next best performing models are \texttt{GPT-text-embedding-ada-002} (prop.~$=71\%$, $d=0.56$) and \texttt{German\_Semantic\_STS\_V2} (prop. $=66\%$, $d=0.60$), followed by \texttt{paraphrase-\allowbreak multilingual-mpnet\allowbreak -base-v2} (prop. $=63\%$, $d=0.36$). The \texttt{Gbert-large}\break (prop.~$=~46\%$, $d=0.04$) and \texttt{G-SciEdBERT} (prop. $=41\%$, $d=-0.09$) models exhibited the weakest performance, with non-significant results.
For the comparison of related concepts based on correctness (RC vs. IRC), \texttt{G-SciEdBERT} emerges as the second-best model (prop.~$=~68\%$, $d=0.40$), followed by \texttt{Gbert-large} (prop. $=57\%$, $d=0.29$) and \texttt{GPT-text\-embedding-ada-002} (prop. $=58\%$, $d=0.27$). In contrast, the lowest-performing models for this comparison are \texttt{German\_Semantic\_STS\_V2} (prop. $=55\%$,\break $d=0.14$) and \texttt{paraphrase\allowbreak -multi\allowbreak lingual-\allowbreak mpnet-base-v2} (prop. $=49\%$,\break $d=-0.03$).

\begin{table}[]
    \centering
    \caption{Results of the instance-based comparisons of similarity values. Abbreviations: prop.~=~proportion of instances for which Eq.~(\ref{eq:sim1}) or (\ref{eq:sim2}) holds; $p$ = $p$-value of a paired one-sided Wilcoxon signed-rank test; $d$ = effect size given by matched-pairs rank biserial correlation coefficient; n.s.: non-significant, *: $p < 0.05$, **: $p < 0.01$, ***: $p < 0.001$.}
    \vspace*{1ex}
    \label{tab3}
    \renewcommand{\arraystretch}{1.2} 
    \begin{tabular}{| m{7.4cm} | >{\centering\arraybackslash}m{0.7cm} | >{\centering\arraybackslash}m{0.7cm} |
                    >{\centering\arraybackslash}m{0.9cm} | > {\centering\arraybackslash}m{0.7cm} 
                    | >{\centering\arraybackslash}m{0.7cm} | >{\centering\arraybackslash}m{0.9cm} | } 
    \hline
        ~  & \multicolumn{3}{|c|}{LT vs. ILT} & \multicolumn{3}{|c|}{RC vs. IRC} \\ 
    \cline{2-4} \cline{5-7}
        Model & prop. & $p$ & $d$ & prop. & $p$ & $d$  \\ 
    \hline
        \texttt{German\_Semantic\_STS\_V2} & 66\% & *** & 0.60 & 55\% & n.s. & 0.14   \\
        \hline
        \texttt{paraph.-multi.-mpnet-base-v2} & 63\% & ** & 0.36 & 49\% & n.s. & -0.03  \\
        \hline
        \texttt{Gbert-large} & 46\% & n.s. & 0.04 & 57\% & ** & 0.29   \\
        \hline
        \texttt{G-SciEdBERT} & 41\% & n.s. & -0.09 & 68\% & ** & 0.40   \\ 
        \hline
        \texttt{GPT-text-embedding-3-large} & 91\% & *** & 0.95 & 83\% & *** & 0.82   \\ 
        \hline
        \texttt{GPT-text-embedding-ada-002} & 71\% & *** & 0.56 & 58\% & * & 0.27   \\ 
    \hline   
    \end{tabular}
\end{table}

\subsection{Evaluation of Embedding Models within a Machine Learning Pipeline}

We compared different embedding models by evaluating their impact on the classification performance (measured by Cohen’s $\kappa$) of a support vector machine (SVM) trained on the model-specific embeddings in categorizing students’ responses into the four categories introduced earlier.

\begin{figure}[]
    \centering
    \Description{The figure contains two vertical bar charts. The left chart is titled "Solely Textual Responses" and the right chart is titled "Responses Including Symbolic Expressions". In both charts, the vertical axis is labeled Cohen’s kappa and ranges from 0 to 1, while the horizontal axis lists six embedding models: German_Semantic_STS_V2, paraphrase-multilingual-mpnet-base-v2, GBERT-large, G-SciEdBERT, GPT-text-embedding-3-large, and GPT-text-embedding-ada-002.

    For textual responses, the kappa values are: German_Semantic_STS_V2 = 0.75, paraphrase-multilingual-mpnet-base-v2 = 0.70, GBERT-large = 0.67, G-SciEdBERT = 0.60, GPT-text-embedding-3-large = 0.82, GPT-text-embedding-ada-002 = 0.76.
    
    For responses with symbolic expressions, the kappa values are: German_Semantic_STS_V2 = 0.59, paraphrase-multilingual-mpnet-base-v2 = 0.59, GBERT-large = 0.60, G-SciEdBERT = 0.52, GPT-text-embedding-3-large = 0.68, GPT-text-embedding-ada-002 = 0.64.}  
    \includegraphics[width=1.00\textwidth]{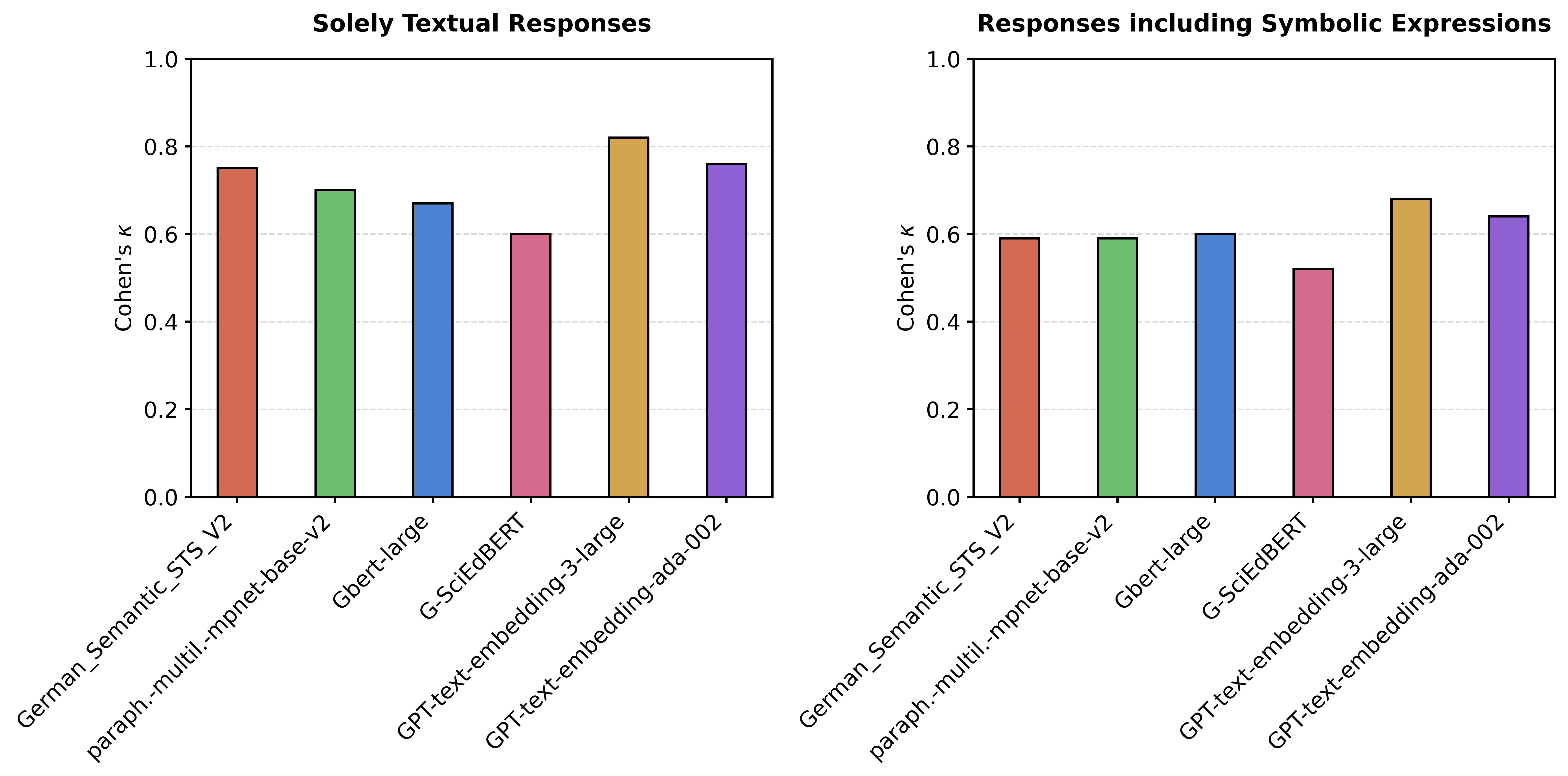}
    \caption{Performance of the SVM classifier, measured by Cohen's $\kappa$, for solely textual student responses (left) and responses including symbolic expressions (right), depending on the NLP embedding model used.}
    \label{fig:klassifikation}
\end{figure}

Figure~\ref{fig:klassifikation} illustrates how Cohen's $\kappa$ depends on the underlying embedding model, in particularly how it differs between solely textual responses and responses including symbolic expressions. As expected, the average classification performance across embedding models is better for solely textual responses ($\bar{\kappa}~=~0.72$) than for responses including symbolic expressions ($\bar{\kappa}~=~0.60$). In more detail, the results indicate that \texttt{German\_Semantic\_STS\_V2}, \texttt{paraphrase-multilingual-mpnet\-base-v2}, and \texttt{Gbert-large} achieved comparable performance for both types of student responses. However, the performance of the \texttt{G-Sci\-EdBERT} model in classifying both types of student responses was significantly lower, consistent with the results presented in Section~\ref{sec:3.1}. The \texttt{GPT-text-embedding-ada-002} model achieved the second-best performance across both response types, while \texttt{GPT-text\-embedding-3-large} performed best, outperforming the weakest embedding model by\break $\Delta\kappa~=~0.22$ for purely textual responses and $\Delta\kappa = 0.16$ for responses including symbolic expressions.

These findings highlight that the choice of embedding model does influence classification performance. While the differences may appear modest at first glance, in large-scale applications such as learning systems, even small improvements can make a meaningful difference. For example, an increase of 0.10 in $\kappa$ can roughly be thought of correctly classifying one additional student response out of every ten—potentially affecting feedback quality for thousands of students in large deployments.

\section{Discussion}

This study revealed substantial differences among NLP embedding models in their capability to process physics-specific symbolic expressions in authentic student responses. Both the similarity-based approach and the more application-oriented machine learning pipeline indicated that some embedding models consistently outperformed others. Notably, OpenAI’s subscription-based model \texttt{GPT-text-embedding-3-large}—which, at the time of the study, was the most recent available—demonstrated superior performance compared to all other models tested, particularly in handling symbolic expressions as well as purely textual input (see Fig.~\ref{fig:klassifikation}, left). This performance advantage compared to non-proprietary models has also been demonstrated in other studies \cite{fengPHYSICSBenchmarkingFoundation2025,petukhovaTextClusteringLarge2025,wangSciBenchEvaluatingCollegelevel2024}, and is likely attributable to the model’s broader training corpus—which presumably includes a greater proportion of symbolic expressions—and to its more advanced embedding architecture compared to the other models. Interestingly, even the \texttt{G-SciEdBERT} embedding model fine-tuned for science education tasks did not appear to effectively handle symbolic expressions, which is likely due to its fine-tuning on more conceptual science tasks rather than tasks emphasizing usage of symbolic expressions \cite{latifGSciEdBERTContextualizedLLM2024}.

These findings have important implications for educational data mining, particularly in educational contexts around science. In science contexts, textual data often contain symbolic expressions that encode valuable diagnostic information. Ignoring such expressions may introduce systematic biases into research results or lead to misguided conclusions and suboptimal decisions in practical applications such as automated assessments. Although differences in embedding model performance may appear modest, our more practical machine learning pipeline approach revealed that \texttt{GPT-text-embedding-3-large} achieved an increase of 0.16 in Cohen’s~$\kappa$ compared to the worst tested embedding model in handling symbolic expressions. Such improvements can have substantial consequences in large-scale applications. For instance, in a hypothetical deployment scenario of the investigated machine learning workflow involving 10,000 student responses, this difference in performance would correspond to approximately 1,600 additional responses being accurately classified.

While NLP embedding models will continue to evolve rapidly, the core insight from this study remains significant: there are substantial and practically relevant differences in how well these models handle symbolic expressions in students' responses. This highlights the importance for both researchers and practitioners to be mindful of these differences when selecting an embedding model, especially in contexts where symbolic expressions may play a central role. However, the choice of an embedding model should not be based solely on performance, as other considerations must be taken into account. One such consideration is cost: while a one-time payment for generating embeddings for a fixed text corpus in a research scenario may be feasible, continuous usage in large-scale applications could result in substantial financial costs \cite{liagkouCostPerspectiveAdopting2024,polveriniMultimodalLargeLanguage2025}. Another critical consideration is regulatory compliance, including data protection and privacy concerns, which may lead institutions or developers to prefer locally hosted models. Additionally, reliance on non-local models, such as those from OpenAI, creates a dependency on external providers. Changes in model architectures, pricing structures, or API terms could impact future applications, posing challenges for the long-term planning of research projects or educational applications.

In light of these aspects, researchers and practitioners concerned with educational data mining and working with textual data that include symbolic expressions must carefully weigh their priorities. Those seeking moderately improved performance in handling symbolic expressions may opt for proprietary models like \texttt{GPT-text-embedding-3-large}, accepting trade-offs such as cost, regulatory constraints, and external dependencies. Alternatively, open-source models may offer greater transparency and control but currently exhibit reduced capabilities in processing symbolic expressions. This trade-off underscores a pressing need for open-source embedding models that are explicitly trained on scientific texts and capable of robustly handling symbolic expressions. In this regard, further research is also needed on explicit approaches \cite{ferreiraMathematicalLanguageProcessing2022}, such as mathematical language processing, that directly aim to understand symbolic expressions rather than treating this capability as a side effect of embedding models primarily designed for general language understanding. 

In the near future, OpenAI can be expected to release its latest generation of embedding models, namely those underlying GPT-4 and potentially GPT-5. These models have demonstrated substantially enhanced capabilities in processing and reasoning with symbolic expressions compared to earlier generations \cite{lopez-simoChallengingChatGPTDifferent2024,tongInvestigatingChatGPT4sPerformance2023,tschisgaleEvaluatingGPTReasoningbased2025a}. It is therefore plausible that the corresponding embedding models will also exhibit improved performance in handling symbolic expressions. However, this hypothesis requires empirical validation once these models become publicly available. We contend that the methodology proposed in this study provides a suitable framework for conducting such an evaluation.

This study has several limitations that also give rise to possible further research directions. 
First, our analysis focused exclusively on physics, a discipline that heavily relies on symbolic expressions. While this focus yields valuable insights, the findings may not be fully generalizable to other scientific domains with distinct symbolic structures, such as chemistry (e.g., molecular formulas) or pure mathematics (e.g., letters representing abstract algebraic objects). Although studies in mathematical contexts already exist \cite{greiner-petterMathwordEmbeddingMath2020,youssefExplorationsUseWord2019}, and initial steps have been taken toward developing open-source embedding models fine-tuned for mathematics education \cite{shenMathBERTPretrainedLanguage2021}, further research is needed that systematically investigates science-specific language and symbolic representations across a broader range of disciplines.

Second, this study did not systematically investigate the robustness of embedding models to variations in notation. For example, formulas can appear in different notational forms (e.g., alternative representations of fractions, summation symbols, or variable names), and it remains unclear how well different embedding models handle this variability. Future research should systematically examine which notation is processed effectively and which presents challenges to better assess model performance and understand their limitations in handling symbolic expressions.

Lastly, the analyses in this study concentrated on symbolic expressions, either in isolation (similarity-based analysis) or with only limited surrounding context (machine learning pipeline analysis). Yet, the embedding models employed are explicitly designed to generate contextualized embeddings and typically perform best when rich contextual information is available, particularly in texts with complex structures, ambiguous word usage, or previously unseen terminology—conditions that are often present in scientific language \cite{aroraContextualEmbeddingsWhen2020}. Consequently, the reported findings regarding the models’ capability to process symbolic expressions should be regarded as a lower-bound estimate of their potential performance. In practical applications where symbolic expressions appear within longer texts, thereby providing richer context, improved performance can reasonably be expected. Nevertheless, this remains an open empirical question, and future research should investigate how contextual information influences the processing of symbolic expressions, including the possibility that additional context may in some cases hinder rather than enhance model performance.

\section{Conclusion}

In summary, this study underscores the importance for educational data mining researchers and practitioners of carefully selecting NLP embedding models for applications involving science-related language products, particularly those containing symbolic expressions such as formulas and equations. Our results reveal meaningful differences in the capabilities of current embedding models to handle physics-specific symbolic expressions, with proprietary models—especially \texttt{GPT-text-embedding-3-large}—\allowbreak demonstrating the best overall performance, though its advantage over other models was moderate rather than decisive. However, even such modest differences can translate into meaningful improvements in large-scale data mining applications, potentially enhancing automated feedback and support for thousands of learners. Nonetheless, embedding model selection must also consider practical factors such as cost, data privacy, and long-term accessibility. Looking ahead, the development of NLP embedding models (or other technologies) that can robustly and transparently process both textual and symbolic information concurrently will be essential for advancing data mining regarding science-related language products that involve symbolic expressions. The evaluation framework presented in this study offers a foundation for systematically benchmarking future embedding models and guiding their application in data mining contexts that involve scientific language.

\section*{Author Contributions}

Both authors contributed equally to the study and the corresponding manuscript. Accordingly, they share first authorship.

\section*{Declaration of Generative AI Software Tools in the Writing Process}

During the preparation of this work, the authors used ChatGPT (OpenAI) throughout the manuscript in order to enhance the clarity, coherence, and style of the text. After using this tool, the authors reviewed and edited the content as needed and take full responsibility for the content of the publication.

\section*{Declaration of Conflicting Interest}

The authors declared no potential conflicts of interest with respect to the research, authorship, and/or publication of this article.

\section*{Funding}

This research was conducted using data from the WinnerS project (supported by the Leibniz Association, Germany under grant no. K194/2015) and data from the LernMINT Research Training Group (funded by the Lower Saxony Ministry of Science and Education under project no. 51410078).


\bibliographystyle{acmtrans}
\bibliography{ref}

\end{document}